\documentclass[letterpaper, 10 pt, conference]{ieeeconf}  %

\IEEEoverridecommandlockouts                              %
\usepackage{array}
\usepackage{url}
\usepackage{amsmath,amssymb}
\usepackage{graphicx}
\newcommand{\distance}{{\textrm{d}}}
\newcommand{\laplace}{{\tilde{\mathcal{L}}}}
\newcommand{\lnear}{{\tilde{L}}}
\newcommand{\event}{{\mathcal{F}}}
\newcommand{\mean}{\mathop{\mathbb{E}}}
\newcommand{\Ic}{\mathcal{I}_c}
\DeclareMathOperator*{\argmin}{arg\,min}
\DeclareMathOperator*{\argmax}{arg\,max}
\newtheorem{thm}{Theorem}[section]

\newtheorem{prop}[thm]{Proposition}

\title{\LARGE \bf
Detecting Structural Shifts in Multivariate Hawkes Processes with Fréchet Statistic
}

\author{Rui Luo and Vikram Krishnamurthy%
\thanks{This work was supported in part by the  U. S. Army Research Office under grant W911NF-21-1-0093, the National Science Foundation under grant CCF-2112457, and City University of Hong Kong under grant 9610639.}%
\thanks{R. Luo is with the Department of Systems Engineering, City University of Hong Kong, Hong Kong SAR, China
        {\tt\small ruiluo@cityu.edu.hk}}%
\thanks{V. Krishnamurthy is with the School of Electrical and Computer Engineering, Cornell University, Ithaca, NY, 14850, USA
        {\tt\small vikramk@cornell.edu}}%
}

\begin{document}

\maketitle
\thispagestyle{empty}
\pagestyle{empty}

\begin{abstract}

This paper proposes a new approach for change point detection in multivariate Hawkes processes using Fréchet statistic of a network. The method splits the point process into overlapping windows, estimates kernel matrices in each window, and reconstructs the signed Laplacians by treating the kernel matrices as the adjacency matrices of the causal network. We demonstrate the effectiveness of our method through experiments on both simulated and cryptocurrency datasets. Our results show that our method is capable of accurately detecting and characterizing changes in the causal structure of multivariate Hawkes processes, and may have potential applications in fields such as finance and neuroscience. The proposed method is an extension of previous work on Fréchet statistics in point process settings and represents an important contribution to the field of change point detection in multivariate point processes.
\end{abstract}

\section{INTRODUCTION}

Networked systems, such as social \cite{luo2022controlling, luo2024conformal}, biological \cite{linderman2016bayesian}, and financial networks \cite{bacry2015hawkes}, often experience events that trigger cascading effects. Social interactions can escalate into conflicts, create polarized groups, or spread misinformation. Neuronal activities in the brain may incite further stimulations or inhibitions. Similarly, initial shocks in stock markets can propagate volatility. These event chains are crucial to understanding the dynamics of practical networks.

Multivariate Hawkes processes are crucial for modeling event occurrences across fields like neuroscience, finance, and social science, effectively capturing self- and cross-excitation among multiple channels that lead to cascading events. Detecting change points in these processes is vital, with current methods primarily focusing on kernel function estimation using generalized likelihood ratio (GLR) or CUSUM procedures. For instance, Yamin et al. \cite{yamin2022online} proposed a modified CUSUM and LRT-based method for supply chain disruptions, while Wang et al. \cite{wang2022sequential} developed a CUSUM procedure for sequential change-point detection in Hawkes networks. Bayesian approaches, such as those by Li et al. \cite{li2020non} and Detommaso \cite{detommaso2019stein}, leverage prior information and tackle complex models. Additionally, Linderman et al. \cite{linderman2014discovering} introduced a random graph model to reveal latent networks in multivariate Hawkes processes. Collectively, these methods advance change point detection research and enhance our understanding of underlying dynamics and future predictions.

In contrast to previous work, we introduce a novel method that simplifies the task of detecting shifts in magnitude and frequency of events by reconstructing the kernel's integral, eliminating the need for full kernel recovery. This method utilizes Fréchet statistics of signed Laplacian matrices to construct test statistics that detect changes, offering a unique network-based perspective on the causal structures driving event dynamics.

Our main contribution is formulating change point detection in multivariate Hawkes processes as a network hypothesis testing problem, enhancing our understanding of underlying causality and providing a robust tool for analyzing multivariate time series data. The efficacy of this approach is validated through its application to practical datasets, effectively identifying significant changes.
The paper is organized as follows: Section \ref{sec:hawkes} outlines the multivariate Hawkes process and defines a kernel matrix that captures the causal network, estimated using the NPHC algorithm in Section \ref{subsec: kernel matrix estimation}. Section \ref{sec:frechet} explores a novel signed Laplacian metric space, enabling the derivation of a computationally efficient Fréchet mean and variance for these causal structures. Our methodology's robustness is demonstrated in Section \ref{sec: empirical} through experiments on simulated data and real cryptocurrency markets, showcasing its effectiveness in pinpointing change points.

\section{Multivariate Hawkes Process}\label{sec:hawkes}
In this section, we outline the formulation of a multivariate Hawkes process. Let $(\Omega, \event, P)$ be a complete probability space and consider a multivariate Hawkes process $N(t)=(N_1, \cdots, N_m)$, where $N_i(t), 1\leq i \leq m$, is the counting process representing the cumulative number of events up to time $t$ for subject $i$. We define a set of increasing $\sigma$-algebras $\{\event^t\}_{t\geq 0}$, where $\event^t = \sigma\{N^t\}$, and the non-negative, $\event^t$-measurable process $\lambda(t)$ as the intensity of $N(t)$, given by:
\begin{equation}
\begin{split}
P(dN_i(t)=1|\event^t) &= \lambda_i(t) dt + o(dt), \\
P(dN_i(t)>1|\event^t) &= o(dt),
\end{split}
\end{equation}
where $o(dt)$ is little-o of $dt$ and $\lim\limits_{dt \rightarrow 0} \frac{o(dt)}{dt} = 0$.
The intensity $\lambda_i(t)$ for each component is given by:
\begin{equation} \label{eq:intensity}
\lambda_i(t) = \mu_i + \sum_{j=1}^{m} \int_{0}^{t} \phi_{ij}(t-s) dN_j(s),
\end{equation}
where $\mu_i$ is the background rate and $\phi_{ij}(\cdot)$ is the kernel function representing the effects of process $N_j$ on process $N_i$. %
A popular choice for $\phi_{ij}(\cdot)$ is an exponential function $\phi_{ij}(t) = \alpha_{ij} e^{-\beta_{ij} t}$, and (\ref{eq:intensity}) can be rewritten as
\begin{equation} \label{eq: exp kernel}
    \lambda_{i}(t) = \mu_i + \sum_{j=1}^{m} \int_{0}^{t} \alpha_{ij} e^{-\beta_{ij} (t-s)} dN_j(s).
\end{equation}

Expressing further in matrix form:
\begin{equation} \label{eq: matrix intensity}
\lambda(t) = \mu + \int_{0}^{t} \boldsymbol{\phi}(t-s) dN(s),
\end{equation}
where $\boldsymbol{\phi}(\cdot)$ is an $m \times m$ kernel matrix with entries $\phi_{ij}(\cdot)$. 
\cite{etesami2016learning} has shown that $\boldsymbol{\phi}(\cdot)$ (which they referred to as the excitation matrix) with exponential kernel functions (\ref{eq: exp kernel}) can reveal the casual structure underlying the multivariate components. 
Specifically, $N_i(t)$ is Granger non-causal for $N_j(t)$ if and only if the corresponding kernel function $\phi_{ij}(\cdot) = 0$ \cite{achab2017uncovering}. %

To infer the causal structure from $\boldsymbol{\phi}(\cdot)$, we use the integrals of its elements to construct the matrix $H=[h_{ij}]$, where
\begin{equation}
h_{ij} = \int_{0}^{\infty} \phi_{ij}(t)dt.
\end{equation}
For kernels with exponential functions (\ref{eq: exp kernel}), we have $h_{ij} = \frac{\alpha_{ij}}{\beta_{ij}}$. 
$h_{ij}$ is commonly referred to as the branching ratio, which quantifies the average number of events triggered by a single event, providing a measure of the endogenous effect in finance \cite{hardiman2014branching}.
Furthermore, the cluster representation of Hawkes processes \cite{hawkes1974cluster} reveals that the integral in the expression for the intensity of a multivariate Hawkes process signifies the average total number of events for subject $i$ that are directly triggered by an event for subject $j$.

\subsection{Kernel Matrix Estimation based on Integrated Cumulants} \label{subsec: kernel matrix estimation}
The Non-Parametric Hawkes with Cumulants (NPHC) algorithm \cite{achab2017uncovering} is a moment-matching method that facilitates the estimation of the kernel matrix $H$ without the need to estimate the kernel functions' shape, which is based on the idea that the integrated cumulants of a Hawkes process can be expressed explicitly as a function of $H$ \cite{jovanovic2015cumulants}.

For an arbitrary $m$-dimensional random vector $X=(X_1, \cdots, X_m)$, the cumulant of order $m$, denoted by $k(X_{\Bar{m}})$, where $\Bar{m}$ indicates the set $\{1, \cdots, m\}$, is defined as
\begin{equation}
    k(X_{\Bar{m}}) = \sum_{\pi} (|\pi|-1)!(-1)^{|\pi|-1} \prod_{B\in \pi} \langle X_B \rangle,
\end{equation}
where the sum goes over all partitions $\pi$ of the set $\{1, \cdots, m\}$, $|\cdot|$ denotes the number of blocks of a given partition, and $\langle X_B \rangle = \langle \prod_{k\in B} X_k \rangle$. 
The integrated cumulants provide a measure of the average correlated activity between events of different subjects, which is a natural generalization of the covariance of two variables to higher dimensions.

Note that if $H$ has a spectral radius $\rho$ strictly smaller than 1, $N(t)$ has asymptotically stationary increments, and $\lambda(t)$ is asymptotically stationary \cite{hawkes1971spectra}. Consequently, $R = (I_d - H)^{-1}$ can be defined.

The first three integrated cumulants of a multivariate Hawkes process can be used to estimate the mean intensity, integrated covariance density matrix, and skewness of the process by equations. 
Specifically, the mean intensity $\Lambda_i$ and the integrated covariance density matrix $C_{ij}$ can be expressed as linear combinations of the vector of background intensities $\mu$ and the matrix $R_{ij}$, while the skewness $K_{ijk}$ can be expressed as a polynomial in $R_{ij}$ and $C_{ij}$ \cite{achab2017uncovering}:
\begin{align}
    \Lambda_i &= \sum_{l=1}^{m} R_{il} \mu_l, \quad
    C_{ij} = \sum_{l=1}^{m} \Lambda_l R_{il} R_{jl}, \\
    K_{ijk} &= \sum_{1=1}^{m} (R_{il}R_{jl}C_{kl} + R_{il}C_{jl}R_{kl} + C_{il}R_{jl}R_{kl} \nonumber \\ 
    & \phantom{-------------} - 2\Lambda_l R_{il}R_{jl}R_{kl}).
\end{align}

The estimation procedure assumes that truncating the integration from $(-\infty, +\infty)$ to $[-W, W]$ introduces only a small error. Given a realization of a stationary Hawkes process $N(t), t\in [0, T]$, where $Z_i$ represents the set of events corresponding to subject $i$, the estimators for the cumulants can be obtained under this assumption:
\begin{align}
    \hat{\Lambda}_i &= \frac{N_i(T)}{T}, \\
    \hat{C}_{ij} &= \frac{1}{T} \sum_{\tau \in Z_i} (N_j(\tau+W) - N_j(\tau-W) - 2W \hat{\Lambda}_j), \label{eq: C_ij hat}\\
    \hat{K}_{ijk} &= \frac{1}{T} \sum_{\tau \in Z_i} \Big[ (N_j(\tau+W) - N_j(\tau-W) - 2W \hat{\Lambda}_j) \nonumber \\
    & \phantom{----} . (N_k(\tau+W) - N_k(\tau-W) - 2W\hat{\Lambda}_k) \Big] \nonumber \\ 
    & \phantom{--} -\frac{\hat{\Lambda}_i}{T} \sum_{\tau\in Z_j} \sum_{\tau' \in Z_k} (2W - |\tau' - \tau|)^+ +  4W^2\hat{\Lambda}_i \hat{\Lambda}_j \hat{\Lambda}_k. \label{eq: K_ijk hat}
\end{align}

Since the covariance $C$ provides fewer independent coefficients than the kernel matrix $H$, the NPHC algorithm focuses on a subset of $m^2$ third-order cumulant coefficients $K_{iij}$, namely $K^c=\{K_{iij}\}_{1\leq i,j \leq m}$. Specifically, the estimator of $R$, denoted $\hat{R}$, is obtained by minimizing the Frobenius norm of two differences:
\begin{align*}
    \hat{R} &= \argmin_R \mathcal{L}(R) \\
    &= \argmin_R  (1-\kappa)\| K^c(R) - \hat{K^c} \|_2^2 + \kappa \| C(R) - \hat{C} \|_2^2,
\end{align*}
where $\|\cdot \|_2$ is the Frobenius norm, while $\hat{C}$ and $\hat{K^c}$ are the respective estimators of $C$ and $K^c$ as defined in (\ref{eq: C_ij hat}), (\ref{eq: K_ijk hat}).
Finally, once $\hat{R}$ is obtained, the kernel matrix $H$ can be estimated as $\hat{H} = I_m - \hat{R}^{-1}$. %

\section{Fréchet Statistics Based Change Point Detection}\label{sec:frechet}
In this section, we propose a change point detection method that utilizes the Fréchet statistics of a sequence of causal networks. We construct these networks using the estimated kernel matrix $\hat{H}$ obtained in Section \ref{subsec: kernel matrix estimation}. By computing the Fréchet distance between the signed Laplacian matrices of these networks, we can detect significant changes in the underlying structure of the multivariate Hawkes process.

\subsection{Dynamic Causal Network}
Our method employs a dynamic signed causal network framework with edges that can have either positive or negative weights. We calculate the node degree by summing the absolute values of the weights on the incoming edges. This degree information is incorporated into the degree matrix, which is crucial in calculating the signed Laplacian for the computation of Fréchet statistics in Section \ref{subsec:metric of graph}.

A causal network is a directed graph where $a_{ij}$ of its weighted adjacency matrix $A$ indicates node $j$'s influence on node $i$, with negative weights for inhibitory influences. 
A dynamic causal network is a sequence of graph snapshots $G^{(1)}, G^{(2)}, \cdots$, where each snapshot $G^{(t)} = (V^{(t)}, E^{(t)})$ represents the causal network observed at time $t$. 
We require that $V^{(1)}=V^{(2)} =\cdots= V$, meaning that the multivariate Hawkes process has a fixed set of components. %

To construct a snapshot of the causal network, we use a sliding window approach. Specifically, we estimate the kernel matrix $\hat{H}$ from a sliding window of the multivariate Hawkes process and take the symmetric part, $A= (\hat{H} + \hat{H}^T)/2$, as the adjacency matrix for the snapshot of the causal network\footnote{In experiments presented in Section \ref{sec: empirical}, we demonstrate that the antisymmetric part of $\hat{H}$, i.e., $(\hat{H} - \hat{H}^T)/2$, is negligible.}. 
Denote $A^{(t)}$ as the adjacency matrix of the causal network during the $t$-th sliding window. The signed Laplacian $\Bar{L}^{(t)}$ is defined as $\Bar{L}^{(t)} = \Bar{D}^{(t)} - A^{(t)}$, where $\Bar{D}^{(t)}$ is the diagonal degree matrix given by $\Bar{d}^{(t)}_{ii} = \sum\limits_{j=1}^{m} |a^{(t)}_{ij}|$, that is, the diagonal entries are the unsigned degree of each node.

\subsection{Fréchet Mean of Signed Laplacian Matrices}\label{subsec:metric of graph}
To measure the dissimilarity between two snapshots in a dynamic causal network, we define a metric space \cite{luo2023frechet} based on the Log-Euclidean metric of the nearest symmetric positive definite (SPD) matrices of their corresponding signed Laplacian matrices. 
The introduced metric enables us to compare the structures of different snapshots of the dynamic causal network. Moreover, it admits a closed-form Fréchet mean, which enables efficient computation. 

The signed Laplacian matrix $\Bar{L}^{(t)}$ is known to be positive semi-definite \cite{kunegis2010spectral}, which can cause problems when using SPD metrics like the Log-Euclidean metric. We address this issue by finding the nearest SPD matrix to $\Bar{L}^{(t)}$ in the Frobenius norm \cite{higham2002computing}. The resulting SPD matrix, denoted by $\tilde{L}^{(t)}$, is used instead of $\Bar{L}^{(t)}$ to define the metric space.

We introduce a metric space $(\laplace, \distance)$ for dynamic causal networks. Here, $\laplace$ denotes the set of nearest SPD matrices to the signed Laplacians, and $\distance$ is a function $\distance,\colon \laplace\times \laplace\to \mathbb {R}_{+}$ defined using the Log-Euclidean metric, i.e., $\delta_{LE}(X, Y) = |\log(X) - \log(Y)|_{\textrm{F}}$.

Suppose we have a set of independent and identically distributed random variables $\lnear^{(1)}, \cdots, \lnear^{(n)} \sim F$ in $(\laplace, \distance)$. According to Theorem 3.13 in \cite{arsigny2007geometric}, the metric space $(\laplace, \distance)$ admits a unique closed-form Fréchet mean $\mu_F$, which is given by $\mu_F = \exp(\mathbb{E}(\log(\lnear)))$. We also define the sample Fréchet mean as $\hat{\mu}_F = \exp(\frac{1}{n} \sum\limits_{i=1}^{n} \log(\lnear^{(i)}))$. The existence and uniqueness of the sample Fréchet mean imply its asymptotic consistency \cite{petersen2019frechet}.

The Fréchet variance quantifies the spread of a random variable around its Fréchet mean. For $\lnear^{(1)}, \cdots, \lnear^{(n)} \sim F$, we define the population Fréchet variance as $V_F = \mathbb{E}(\distance^2(\mu_F, \lnear))$ and the sample Fréchet variance as $\hat{V}_F = \frac{1}{n} \sum\limits_{i=1}^{n} \distance^2(\hat{\mu}_F, \lnear^{(i)}) = \frac{1}{n} \sum\limits_{i=1}^{n} |\frac{1}{n} \sum\limits_{j=1}^{n} \log(\lnear^{(j)}) - \log(\lnear^{(i)}) |_{\textrm{F}}^2$.

We establish the Central Limit Theorem (CLT) for the sample Fréchet variance $\hat{V}_F$ in the metric space $(\laplace, \distance)$ as follows: 
\begin{prop}[Fréchet Variance CLT]\label{prop:CLT variance}
    Under the following assumptions (Assumptions 1-3 in \cite{dubey2019frechet}):
    \begin{enumerate}
        \item $\mu_F$ and $\hat{\mu}_F$ exist and are unique;
        \item For any $l \in \laplace$, $\delta J(\delta, l) \rightarrow 0$ as $\delta \rightarrow 0$, where the complexity of the metric space $\laplace$ is quantified by a bound on the entropy integral for metric $\delta$-balls $B_{\delta}(l)$:
        \begin{equation}
            J(\delta, l) = \int_{0}^{1} [1 + \log N\{\epsilon \delta / 2, B_{\delta}(l), \distance \}]^{1/2} d\epsilon, \nonumber
        \end{equation}
        where $B_{\delta}(l)$ is the $\delta$-ball centered at $l$ in the metric $\distance$ and $N\{\epsilon \delta / 2, B_{\delta}(l), \distance \}$ is the covering number for $B_{\delta}(l)$ using open balls of radius $\epsilon \delta/2$.
        
        \item The entropy integral of the metric space is finite, i.e., $\int_{0}^{1} [1 + \log N(\epsilon, \laplace, \distance)]^{1/2} d\epsilon < \infty$.
    \end{enumerate}
    The following result is obtained:
    \begin{equation}\label{eq:CLT variance}
    \begin{split}
        n^{1/2} (\hat{V}_F - V_F) \rightarrow N(0, \sigma^2_F) \; \textrm{in distribution,}
    \end{split}
    \end{equation}
where $\sigma^2_F = \textrm{Var}\{\distance^2(\mu_F, \lnear) \}$.
\end{prop}

\subsection{Estimating the Location of a Change Point}\label{subsec:one change point}

\subsubsection{Test Statistics Construction}
Let us consider a sequence $\{Y^{(i)}\}_{i=1}^{n}$ of independent, time-ordered data points in a metric space $(\laplace, \distance)$ defined in Section \ref{subsec:metric of graph}. We assume there is at most one change point, which we denote by $0<\tau<1$. Specifically, $Y^{(1)}, \cdots, Y^{(\lfloor n\tau \rfloor)} \sim F_1$ and $Y^{(\lfloor n\tau \rfloor +1)}, \cdots, Y^{(n)} \sim F_2$, where $F_1$ and $F_2$ are unknown probability measures on $(\laplace, \distance)$, and $\lfloor x \rfloor$ is the greatest integer less than or equal to $x$. Our aim is to test the null hypothesis of distribution homogeneity, denoted by $H_0: F_1 = F_2$, against the alternative hypothesis of a single change point, denoted by $H_1: F_1 \neq F_2$.

We constrain the hypothesized change point location $\tau$ to a compact interval $\Ic=[c, 1-c] \subset [0, 1]$ with positive constant $c$ to ensure accurate representation of each segment's Fréchet mean and variance. For statistical analysis of segments separated by $u \in \Ic$, we compute the sample Fréchet mean before and after $\lfloor nu \rfloor$ observations as:
\begin{equation}
\begin{split}
\hat{\mu}_{[0, u]} &= \argmin\limits_{l \in \laplace} \frac{1}{\lfloor n\tau \rfloor} \sum\limits_{i=1}^{\lfloor n\tau \rfloor} \distance^2(Y^{(i)}, l),\\
\hat{\mu}_{[u, 1]} &= \argmin\limits_{l \in \laplace} \frac{1}{n-\lfloor n\tau \rfloor} \sum\limits_{i=\lfloor n\tau \rfloor +1}^{n} \distance^2(Y^{(i)}, l), \nonumber
\end{split}
\end{equation}
and the corresponding sample Fréchet variances are:
\begin{equation}\label{eq:variance}
\begin{split}
\hat{V}_{[0, u]} &= \frac{1}{\lfloor n\tau \rfloor} \sum\limits_{i=1}^{\lfloor n\tau \rfloor} \distance^2(Y^{(i)}, \hat{\mu}_{[0, u]}),\\
\hat{V}_{[u, 1]} &= \frac{1}{n-\lfloor n\tau \rfloor} \sum\limits_{i=\lfloor n\tau \rfloor +1}^{n} \distance^2(Y^{(i)}, \hat{\mu}_{[u, 1]}),
\end{split}
\end{equation}

One can obtain the contaminated version of Fréchet variances by replacing the Fréchet mean of a segment with the mean of the complementary segment. This leads to the definitions:
\begin{equation}
\begin{split}
\hat{V}^C_{[0, u]} &= \frac{1}{\lfloor n\tau \rfloor} \sum\limits_{i=1}^{\lfloor n\tau \rfloor} \distance^2(Y^{(i)}, \hat{\mu}_{[u, 1]}),\\
\hat{V}^C_{[u, 1]} &= \frac{1}{n-\lfloor n\tau \rfloor} \sum\limits_{i=\lfloor n\tau \rfloor +1}^{n} \distance^2(Y^{(i)}, \hat{\mu}_{[0, u]}), \nonumber
\end{split}
\end{equation}
which are guaranteed to be at least as large as the correct version (\ref{eq:variance}). The differences $\hat{V}^C_1 - \hat{V}_{[0, u]}$ and $\hat{V}^C_2 - \hat{V}_{[u, 1]}$ can be interpreted as measures of the between-group variance of the two segments.

For some fixed $u \in \Ic$, the statistic $\sqrt{u(1-u)}(\sqrt{n}/\sigma)(\hat{V}_{[0, u]} - \hat{V}_{[u, 1]})$ has an asymptotic standard normal distribution under the null hypothesis $H_0$. Here, $\sigma$ denotes the asymptotic variance of the empirical Fréchet variance. This result allows us to test hypotheses about differences in Fréchet variances between two segments of data. A sample based estimator for $\sigma^2$ is:
\begin{equation}
\hat{\sigma}^2 = \frac{1}{n} \sum_{i=1}^{n} \distance^4(\hat{\mu}, \lnear_i) - \left( \frac{1}{n} \sum_{i=1}^{n} \distance^2(\hat{\mu}, \lnear_i) \right)^2, \nonumber
\end{equation}
which is consistent under $H_0$ \cite{dubey2019frechet}. Also,
\begin{equation}
\hat{\mu}=\argmin\limits_{l\in \laplace} \frac{1}{n} \sum\limits_{i=1}^{n} \distance^2(Y^{(i)}, l), \quad
\hat{V} = \frac{1}{n} \sum\limits_{i=1}^{n} \distance^2(Y^{(i)}, \hat{\mu}). \nonumber
\end{equation}

The test statistic proposed in \cite{dubey2020frechet} can detect differences in both Fréchet means and Fréchet variances of the distributions $F_1$ and $F_2$. The equation for the test statistic is provided as follows:
\begin{equation}\label{eq:Tn(u)}
\begin{split}
T_n(u) = & \frac{u(1-u)}{\hat{\sigma}^2} \Big[ (\hat{V}_{[0, u]} - \hat{V}_{[u, 1]})^2 \\
& \phantom{---} + (\hat{V}_{[0, u]}^C - \hat{V}_{[0, u]} + \hat{V}_{[u, 1]}^C - \hat{V}_{[u, 1]})^2 \Big]
\end{split}
\end{equation}
In this equation, $(\hat{V}_{[0, u]} - \hat{V}_{[u, 1]})^2$ indicates the difference in Fréchet variances between two segments of data, while $(\hat{V}_{[0, u]}^C - \hat{V}_{[0, u]} + \hat{V}_{[u, 1]}^C - \hat{V}_{[u, 1]})^2$ captures the difference in Fréchet means between the two segments. 

Under $H_0$ that no change point exists, we establish the weak convergence\footnote{Weak convergence is a function space generalization of convergence in distribution \cite{billingsley2013convergence}.} of $nT_n(u)$:

\begin{prop}[Weak Convergence of $nT_n(u)$]\label{prop:weak convergence}
    Under $H_0$ and the following assumptions (Assumptions 1-4 in \cite{dubey2020frechet}):
    \begin{enumerate}
        \item $\mu_F$ and $\hat{\mu}_F$ exist and are unique;
        \item For any $\alpha=\{\alpha_1, \dots, \alpha_n: 0\leq \alpha_i \leq 1, \sum_{i=1}^{n} \alpha_i = 1\}$, $\hat{\mu}_{\alpha}=\argmin_{l\in \laplace} \sum_{i=1}^{n} \alpha_i \distance^2(Y^{(i)}, l)$ exists and is unique almost surely;
        \item For any $l \in \laplace$, $\int_{0}^{1} \sqrt{\log N(\epsilon \delta, B_{\delta}(l), \distance)} d\epsilon = O(1)$ as $\delta \rightarrow 0$;
        \item There exist $\delta>0$ and $C>0$ such that for all $l \in B_{\delta}(\mu_F)$:
        $\mean(\distance^2(Y, l)) - \mean(\distance^2(Y, \mu_F)) = C \distance^2(l, \mu_F) + O(\delta^2) \textrm{ as $\delta \rightarrow 0$}$,
    \end{enumerate}
    The following result is obtained:    \begin{equation}\label{eq:weak convergence}
        \{nT_n(u): u\in \Ic \} \Rightarrow \{ \mathcal{G}^2(u): u\in \Ic \},
    \end{equation}
    where $\mathcal{G} = \left\{ \frac{\mathcal{B}(u)}{\sqrt{u(1-u)}}: u\in \Ic \right\}$ and $\mathcal{B}(u)$ is a Brownian bridge on $\Ic$, which is a Gaussian process indexed by $\Ic$ with zero mean and covariance structure given by $K(s, t) = \min(s,t) - st$.
\end{prop}

To perform a hypothesis test between $H_0$ and $H_1$, the statistic $\sup_{u\in \Ic} nT_n(u)$ is used. Here, $T_n(u)$ is a test statistic for hypothesis testing, calculated for each potential change point $u\in \Ic$. The $(1-\alpha)$th quantile of $\sup_{u\in \Ic} \mathcal{G}^2(u)$ is denoted as $q_{1-\alpha}$, which is obtained by a bootstrap approach, as described in Section 3.3 of \cite{dubey2020frechet}.

Under the null hypothesis $H_0$: 
\begin{equation}
\sup_{u \in \Ic} nT_n(u) \Rightarrow \sup_{u \in \Ic} \mathcal{G}^2(u),
\end{equation}
where $T_n(u)$ is the test statistic and $\mathcal{G}^2(u)$ is the limiting distribution. We use this result to define the rejection region for a level $\alpha$ significance test as:
\begin{equation}
R_{n,\alpha} = \left\{ \sup_{u \in \Ic} nT_n(u) > q_{1-\alpha} \right\},
\end{equation}
where $q_{1-\alpha}$ is the $(1-\alpha)$ quantile of $\mathcal{G}^2(u)$.

Under the alternative hypothesis $H_1$, which assumes a change point is present at $\tau \in \Ic$, we can locate it by finding the maximizer of the process $T_n(u)$:
\begin{equation}
\hat{\tau} = \argmax_{u \in \Ic} T_n(u) = \argmax_{\lfloor nc \rfloor \leq k \leq n - \lfloor nc \rfloor} T_n\left(\frac{k}{n}\right),
\end{equation}
where $\hat{\tau}$ is the estimated change point that maximizes the test statistic across all potential change points. 
Furthermore, we propose a binary segmentation procedure that extends the proposed statistic $T_n(u)$ to multiple change point scenario \cite{luo2023frechet}.

\begin{figure}
	\centering
	\includegraphics[width=0.5\textwidth]{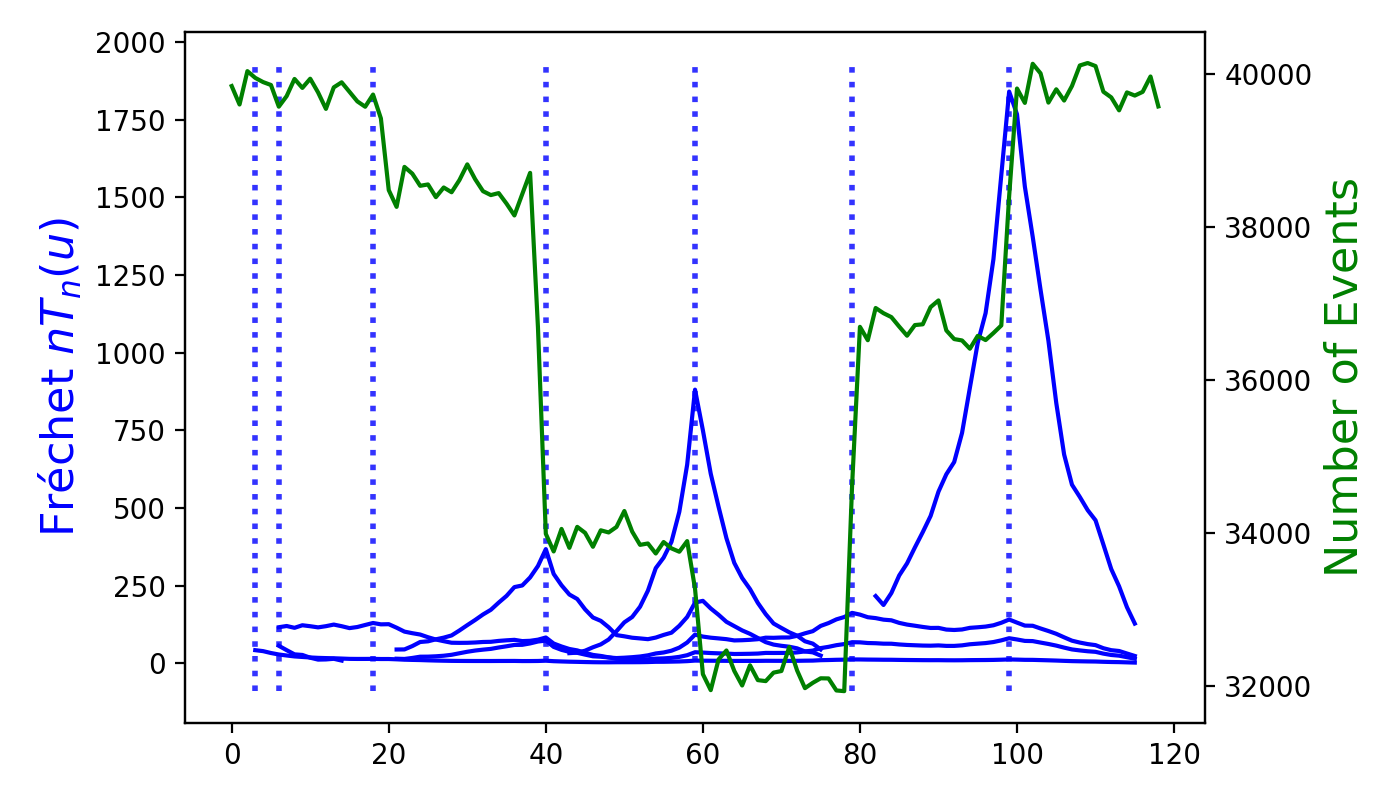}
	\caption{Our algorithm effectively detected all change points in a synthetic 10-dimensional point process of length 600000 using Fréchet statistics. The data was split into 119 overlapping windows of length 10000, with the number of events in each window shown in green. The Fréchet test statistic ${nT_n(u): u\in \Ic}$ was also plotted for each analyzed segment due to the use of binary segmentation. The accuracy of our algorithm was validated by the ground truth, as the peak of the test statistics occurs around the significant changes in the number of events. }
	\label{fig: synthetic}
\end{figure}

\section{Empirical Analysis on Simulated and Practical Networks}\label{sec: empirical}
We evaluate the performance of the proposed method on change point detection on both a simulated dataset and a cryptocurrency market dataset. Our results are completely reproducible; the code and datasets used in the experiments are publicly available at \url{https://tinyurl.com/Hawkes-CPD}.

\subsection{Simulation Study}
We generated a synthetic dataset using the multivariate Hawkes process model, implemented through the Python library \texttt{tick}\footnote{\url{https://github.com/X-DataInitiative/tick}}. We first set the base intensities $\lambda_i$ for all 10 subjects to be 0.3. We then defined the kernel function using an exponential decay model, as shown in (\ref{eq: exp kernel}), with decay parameters $\beta_{ij} = 0.5 + u_{ij}$, where $u_{ij}$ is sampled from a uniform distribution between 0 and 1.

To set the amplitude of the kernel function, we used the following values: for $i=j$, $\hat{\alpha}_{ij}=1/8$, and for $i\neq j$, $\hat{\alpha}_{ij}$ was sampled from a normal distribution with mean 0 and variance $1/64$. We symmetrized these values to obtain $\alpha_{ij} = (\hat{\alpha}_{ij} + \hat{\alpha}_{ji})/2$. We ensured that the spectral radius of the resulting kernel matrix $H$ was less than 1, which guarantees the stationarity of the process. The integral of each kernel function is computed as $h_{ij} = \frac{\alpha_{ij}}{\beta_{ij}}$.

We induce change points by updating the $\alpha_{ij}$ every 100000 units, resulting in a modification of the causal network $H$ of the process.

To evaluate the performance of our algorithm in detecting multiple change points, we analyze a simulated multivariate process of length 600000. The data is divided into overlapping windows of length 10000, with adjacent windows having an overlapping window of length 5000. %
Figure \ref{fig: synthetic} shows that the Fréchet statistics based algorithms accurately detect all the change points, as confirmed by the ground truth generated during dataset construction. We also plot the test statistics, that is, the Fréchet test statistic ${nT_n(u): u\in \Ic}$ (\ref{eq:Tn(u)}) for our algorithm. Since we use binary segmentation, we display the Fréchet test statistic for each analyzed segment.

\subsection{Cryptocurrency Price Data}
We use Binance's cryptocurrency price data\footnote{\url{https://github.com/binance-us/binance-official-api-docs}} for 41 DeFi-related coins. %
The data was sampled every 5 minutes for 396 days between 1st March 2022 and 31st March 2023. Events were identified whenever a coin price changed by $\pm 0.5\%$ of its current price, resulting in a 41-dimensional multivariate point process. We divided the process into 56 overlapping 2-week windows with 1-week overlap for adjacent windows.

\begin{figure}
	\centering
	\includegraphics[width=0.5\textwidth]{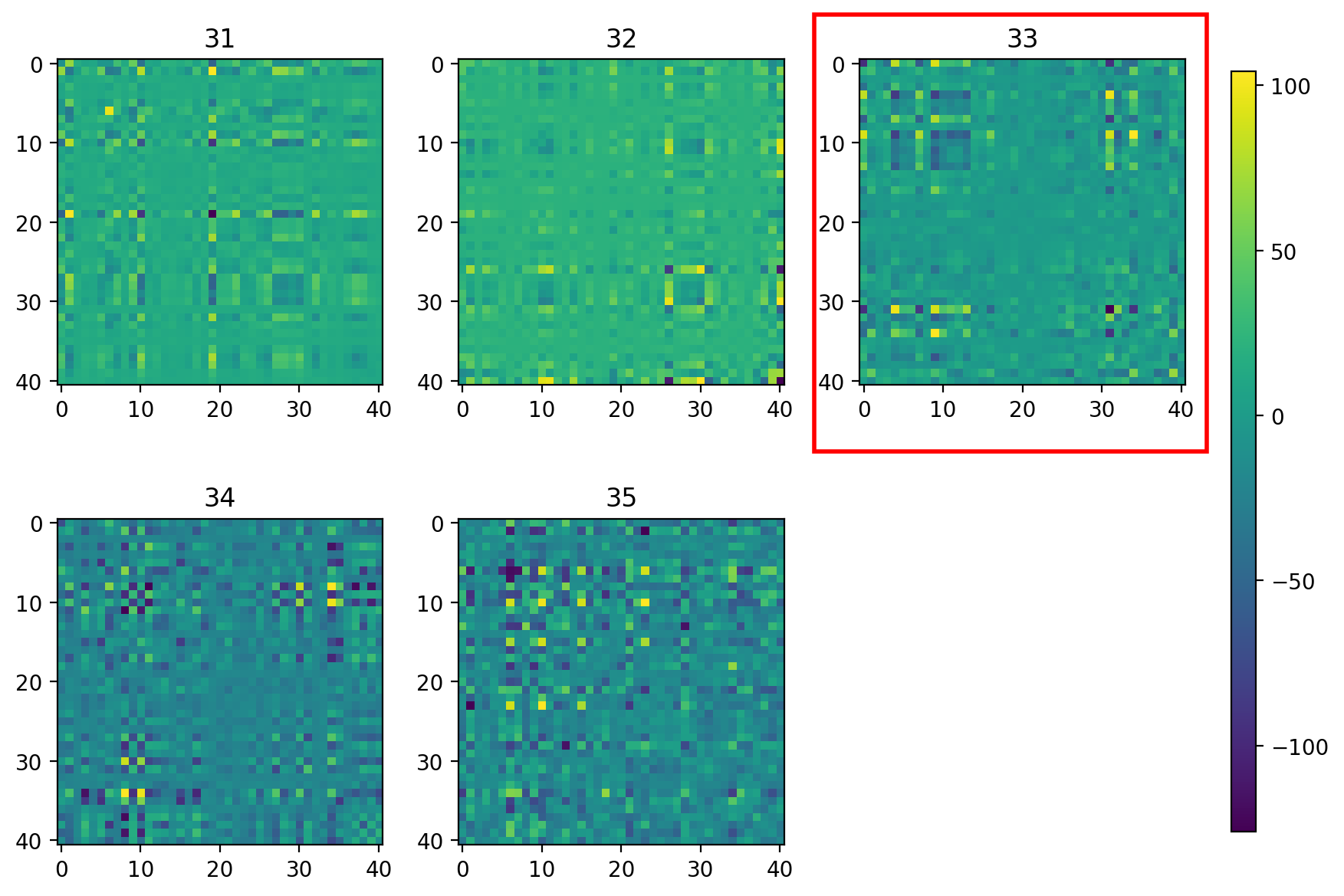}
	\caption{The sequence of symmetrized kernel matrices $A^{(t)}=(H^{(t)} + {H^{(t)}}^T)/2$ estimated by the NPHC algorithm (see Section \ref{subsec: kernel matrix estimation}) is displayed before and after the detected change point at time 33. The change in the underlying causal network dynamics is reflected in the shift of the kernel matrix patterns from the earlier (left) to the later (right) time periods. }
	\label{fig: adj}
\end{figure}

\begin{table}\label{table:enron}
\begin{tabular}{ | m{5em} | m{0.6cm}| m{5.2cm} | } 
\hline
    Start Date & No. & Event  \\ \hline 
   06/06/2022& 14  & Terra-LUNA contagion. \\ \hline 
   07/11/2022&  19 & Voyager and Celsius filed for bankruptcy. \\ \hline 
   10/17/2022&  33 & Over US\$718 million stolen from decentralized finance\footnote{\url{https://forkast.news/october-biggest-month-biggest-year-crypto-hacking/}}. \\ \hline
   12/05/2022&  40 & BlockFi, which had taken a US\$250 million loan from FTX, declared bankruptcy. \\ \hline
   01/16/2023&  46 & Bitcoin price is up 39\% since the start of January. \\ \hline
\end{tabular}
    \caption{Timeline of cryptocurrency market events.}
\end{table}

\begin{figure}
	\centering
	\includegraphics[width=0.5\textwidth]{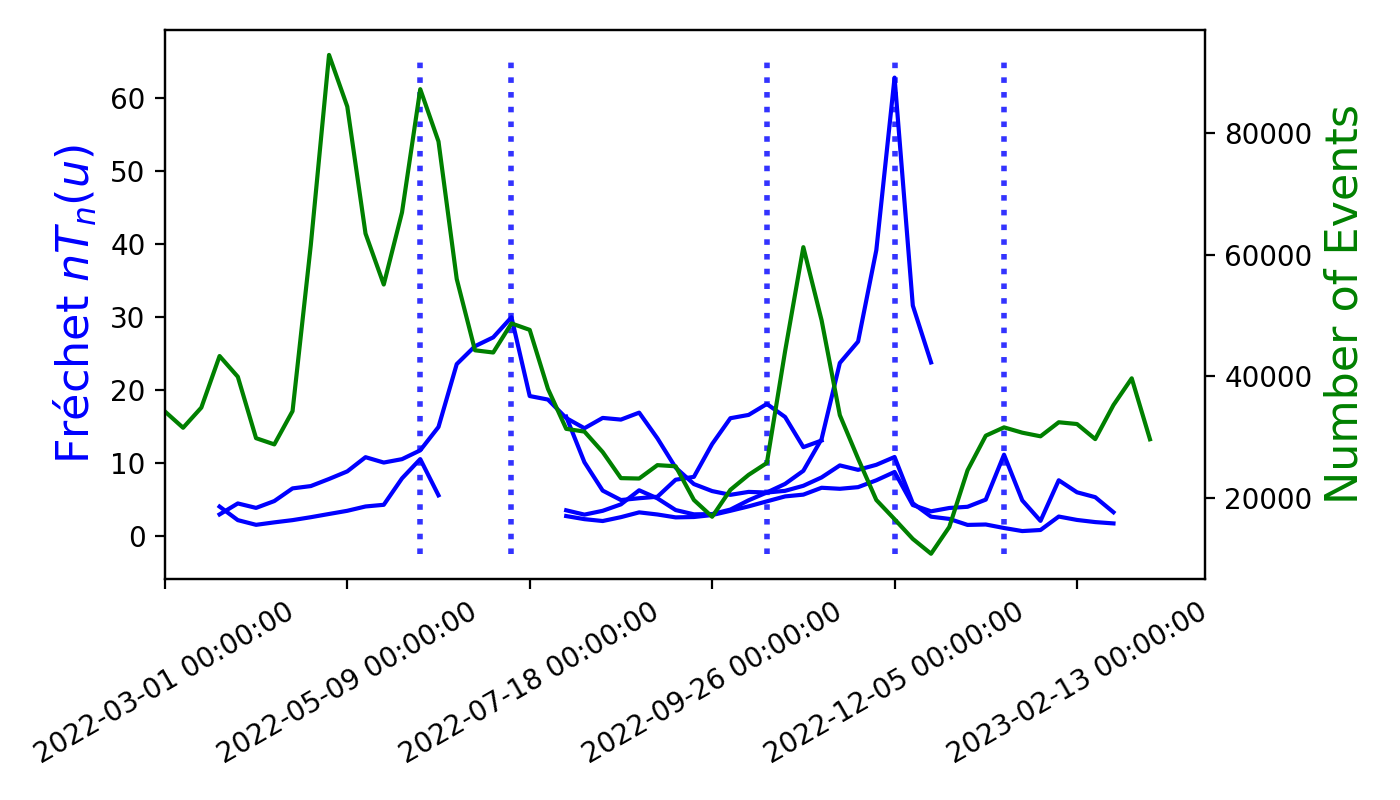}
	\caption{Multiple change point detection results of the cryptocurrency price dataset. The number of events in each window is displayed in green.}
	\label{fig: crypto}
\end{figure}

We present the results of our analysis of the cryptocurrency price dataset using a 41-dimensional multivariate Hawkes process model. Figure \ref{fig: crypto} displays the multiple change points detected in the model, which highlight periods of increased activity and volatility in the cryptocurrency market. To better understand the underlying dynamics, we show the adjacency matrices of the causal network of the multivariate Hawkes process before and after one of the detected change points in Figure \ref{fig: adj}.

To validate the detected change points, we cross-referenced them with significant events in the cryptocurrency market, as summarized in Table 1. Our findings offer valuable insights into the communication patterns and organizational behavior during market transformations or even crashes, and demonstrate the potential of multivariate Hawkes process models in capturing the complex dynamics of financial markets.

\addtolength{\textheight}{15cm}   %

\section{Conclusions}
\label{sec:conclusion}
In this study, we propose a method for detecting change points in causal networks of multivariate Hawkes processes using Fréchet statistics. Specifically, we divide the point process into a sequence of overlapping windows, estimate the kernel matrices in each window, and reconstruct the signed Laplacians by treating the kernel matrices as the graph adjacency matrices of the causal network. We evaluate the method on both simulated and practical datasets, which demonstrate that the proposed method is a powerful tool for detecting and characterizing changes in the causal structure of multivariate Hawkes processes.


\begin{thebibliography}{10}

\bibitem{luo2022controlling}
R. Luo, B. Nettasinghe, and V. Krishnamurthy,
\newblock ``Controlling segregation in social network dynamics as an edge formation game,''
\newblock {\em IEEE transactions on network science and engineering}, vol. 9, no. 4, pp. 2317--2329, 2022.

\bibitem{luo2024conformal}
R. Luo and N. Colombo,
\newblock ``Conformal load prediction with transductive graph autoencoders,''
\newblock {\em arXiv preprint arXiv:2406.08281}, 2024.

\bibitem{linderman2016bayesian}
S.~W. Linderman,
\newblock {\em Bayesian Methods for Discovering Structure in Neural Spike Trains},
\newblock Ph.D. thesis, Harvard University, 2016.

\bibitem{bacry2015hawkes}
E. Bacry, I. Mastromatteo, and J.-F. Muzy,
\newblock ``Hawkes processes in finance,''
\newblock {\em Market Microstructure and Liquidity}, vol. 1, no. 01, pp. 1550005, 2015.

\bibitem{yamin2022online}
K. Yamin, H. Wang, B. Montreuil, and Y. Xie,
\newblock ``Online detection of supply chain network disruptions using sequential change-point detection for hawkes processes,''
\newblock {\em arXiv preprint arXiv:2211.12091}, 2022.

\bibitem{wang2022sequential}
H. Wang, L. Xie, Y. Xie, A. Cuozzo, and S. Mak,
\newblock ``Sequential change-point detection for mutually exciting point processes,''
\newblock {\em Technometrics}, pp. 1--13, 2022.

\bibitem{li2020non}
Q. Li, F. Guo, and I. Kim,
\newblock ``A non-parametric bayesian change-point method for recurrent events,''
\newblock {\em Journal of Statistical Computation and Simulation}, vol. 90, no. 16, pp. 2929--2948, 2020.

\bibitem{detommaso2019stein}
G. Detommaso, H. Hoitzing, T. Cui, and A. Alamir,
\newblock ``Stein variational online changepoint detection with applications to hawkes processes and neural networks,''
\newblock {\em arXiv preprint arXiv:1901.07987}, 2019.

\bibitem{linderman2014discovering}
S. Linderman and R. Adams,
\newblock ``Discovering latent network structure in point process data,''
\newblock in {\em International conference on machine learning}. PMLR, 2014, pp. 1413--1421.

\bibitem{etesami2016learning}
J. Etesami, N. Kiyavash, K. Zhang, and K. Singhal,
\newblock ``Learning network of multivariate hawkes processes: A time series approach,''
\newblock {\em arXiv preprint arXiv:1603.04319}, 2016.

\bibitem{achab2017uncovering}
M. Achab, E. Bacry, S. Ga{\i}ffas, I. Mastromatteo, and J.-F. Muzy,
\newblock ``Uncovering causality from multivariate hawkes integrated cumulants,''
\newblock in {\em International Conference on Machine Learning}. PMLR, 2017, pp. 1--10.

\bibitem{hardiman2014branching}
S.~J. Hardiman and J.-P. Bouchaud,
\newblock ``Branching-ratio approximation for the self-exciting hawkes process,''
\newblock {\em Physical Review E}, vol. 90, no. 6, pp. 062807, 2014.

\bibitem{hawkes1974cluster}
A.~G. Hawkes and D. Oakes,
\newblock ``A cluster process representation of a self-exciting process,''
\newblock {\em Journal of applied probability}, vol. 11, no. 3, pp. 493--503, 1974.

\bibitem{jovanovic2015cumulants}
S. Jovanovi{\'c}, J. Hertz, and S. Rotter,
\newblock ``Cumulants of hawkes point processes,''
\newblock {\em Physical Review E}, vol. 91, no. 4, pp. 042802, 2015.

\bibitem{hawkes1971spectra}
A.~G. Hawkes,
\newblock ``Spectra of some self-exciting and mutually exciting point processes,''
\newblock {\em Biometrika}, vol. 58, no. 1, pp. 83--90, 1971.

\bibitem{luo2023frechet}
R. Luo and V. Krishnamurthy,
\newblock ``Fréchet-statistics-based change point detection in dynamic social networks,''
\newblock {\em IEEE Transactions on Computational Social Systems}, pp. 1--9, 2023.

\bibitem{kunegis2010spectral}
J. Kunegis, S. Schmidt, A. Lommatzsch, J. Lerner, E.~W. De~Luca, and S. Albayrak,
\newblock ``Spectral analysis of signed graphs for clustering, prediction and visualization,''
\newblock in {\em Proceedings of the 2010 SIAM international conference on data mining}. SIAM, 2010, pp. 559--570.

\bibitem{higham2002computing}
N.~J. Higham,
\newblock ``Computing the nearest correlation matrix—a problem from finance,''
\newblock {\em IMA journal of Numerical Analysis}, vol. 22, no. 3, pp. 329--343, 2002.

\bibitem{arsigny2007geometric}
V. Arsigny, P. Fillard, X. Pennec, and N. Ayache,
\newblock ``Geometric means in a novel vector space structure on symmetric positive-definite matrices,''
\newblock {\em SIAM journal on matrix analysis and applications}, vol. 29, no. 1, pp. 328--347, 2007.

\bibitem{petersen2019frechet}
A. Petersen and H.-G. M{\"u}ller,
\newblock ``{Fréchet regression for random objects with Euclidean predictors},''
\newblock {\em The Annals of Statistics}, vol. 47, no. 2, pp. 691 -- 719, 2019.

\bibitem{dubey2019frechet}
P. Dubey and H.-G. M{\"u}ller,
\newblock ``Fr{\'e}chet analysis of variance for random objects,''
\newblock {\em Biometrika}, vol. 106, no. 4, pp. 803--821, 2019.

\bibitem{dubey2020frechet}
P. Dubey and H.-G. M{\"u}ller,
\newblock ``{Fréchet change-point detection},''
\newblock {\em The Annals of Statistics}, vol. 48, no. 6, pp. 3312 -- 3335, 2020.

\bibitem{billingsley2013convergence}
P. Billingsley,
\newblock {\em Convergence of probability measures},
\newblock John Wiley \& Sons, 2013.

\end{thebibliography}
\end{document}